\documentclass[journal]{IEEEtai}

\usepackage[colorlinks,urlcolor=blue,linkcolor=blue,citecolor=blue]{hyperref}

\usepackage{color,array}

\usepackage{graphicx}
\usepackage{algorithm}
\usepackage{algpseudocode}
\usepackage{multirow}
\usepackage{graphicx}
\usepackage{amsmath}
\usepackage{booktabs}
\usepackage{amssymb}
\usepackage{color}
\usepackage{booktabs}
\usepackage{autobreak}
\usepackage{subfig}
\usepackage{pifont}
\usepackage{makecell}
\usepackage{bbding}
\usepackage{graphicx}
\usepackage{float}
\usepackage{booktabs}
\usepackage{makecell}
\usepackage{multirow}
\usepackage{caption}
\usepackage{graphicx}
\usepackage{amssymb}
\usepackage{amsmath}
\usepackage{subcaption}
\usepackage{float}
\usepackage{booktabs}
\usepackage{ulem}
\usepackage{multirow}
\usepackage[table]{xcolor}
\usepackage{arydshln}

\begin{document}

\title{Single-Teacher View Augmentation: Enhancing Knowledge Distillation with Student-Guided Perturbations} 

\author{Xuyi~Yu, Yaohua Liu, Chengjun~Li, Qiang~Tang,  Shuzhe~Tang, and Kuizhi~Mei,~\IEEEmembership{Member,~IEEE}

\thanks{Xuyi~Yu, Shuzhe~Tang and Kuizhi~Mei are with the State Key Laboratory of Human-Machine Hybrid Augmented Intelligence, Institute of Artificial Intelligence and Robotics, Xi'an Jiaotong University, Xi'an 710049, China (e-mail: yuxuyi@stu.xjtu.edu.cn; 2223315783@stu.xjtu.edu.cn; meikuizhi@mail.xjtu.edu.cn).}

\thanks{Yaohua Liu is with the Chinese Medicine Guangdong Laboratory (Hengqin Laboratory), Zhuhai 519031, Guangdong, China (e-mail: liuyaohua@hqcmlab.cn).}

\thanks{Chengjun~Li and Qiang~Tang are with the Shanghai ZEEKR Blue New Energy Technology CO., Ltd., Shanghai, China (e-mail: Burt.Lee@zeekrlife.com; Qiang.Tang2@zeekrlife.com).}}


\maketitle

\begin{abstract}
Knowledge distillation (KD) typically relies on the fixed perspective of 
a single teacher, limiting the diversity of supervisory signals. While 
multi-teacher distillation addresses this by aggregating knowledge from 
multiple models, it incurs prohibitive computational and storage costs. 
To balance efficiency and diversity, recent research has focused on 
generating virtual views from a single teacher. However, existing methods 
face a trade-off: random perturbation approaches offer efficiency but 
lack controlled diversity, while structured augmentation methods require 
multi-stage training and incur linear parameter growth. 
We observe that this trade-off stems from a common design choice: 
using the teacher's strong but static features to generate views. 
Instead, we propose Shift-Augmented Knowledge Distillation (SAKD), 
a simple yet effective framework that leverages the student's evolving 
features as a dynamic condition for perturbation generation. This 
shift in perspective enables single-stage training while producing 
adaptive, diverse views through a parameter-free cyclic shift. 
Extensive experiments on CIFAR-100 and ImageNet demonstrate that 
SAKD consistently outperforms random perturbation methods and 
achieves accuracy on par with two-stage approaches, while using 
significantly fewer parameters and eliminating pre-training 
requirements.
\end{abstract}



\section{Introduction}
\label{sec:introduction}

The deployment of deep neural networks in edge devices, mobile platforms, and other resource-constrained environments is often hindered by their substantial computational and memory requirements. Knowledge distillation (KD) \cite{hinton2015distilling} has emerged as a leading solution, transferring knowledge from a cumbersome but accurate teacher model to a lightweight student. While original KD aligned softened output probabilities, subsequent research has expanded to richer supervisory forms including feature-based \cite{zagoruyko2016paying,tian2019contrastive,ahn2019variational,chen2021distilling}, relation-based \cite{park2019relational,peng2019correlation,guo2023class}, and logit-based approaches \cite{zhao2022decoupled,jin2023multi,guo2020reducing,li2023curriculum,sun2024logit,tai1,tai2}. Despite these advances, a critical limitation remains: the student receives supervision from only one fixed perspective of the teacher, limiting the diversity of guidance and potentially constraining generalization.

\begin{figure*}[t]
    \centering
    \includegraphics[width=\linewidth]{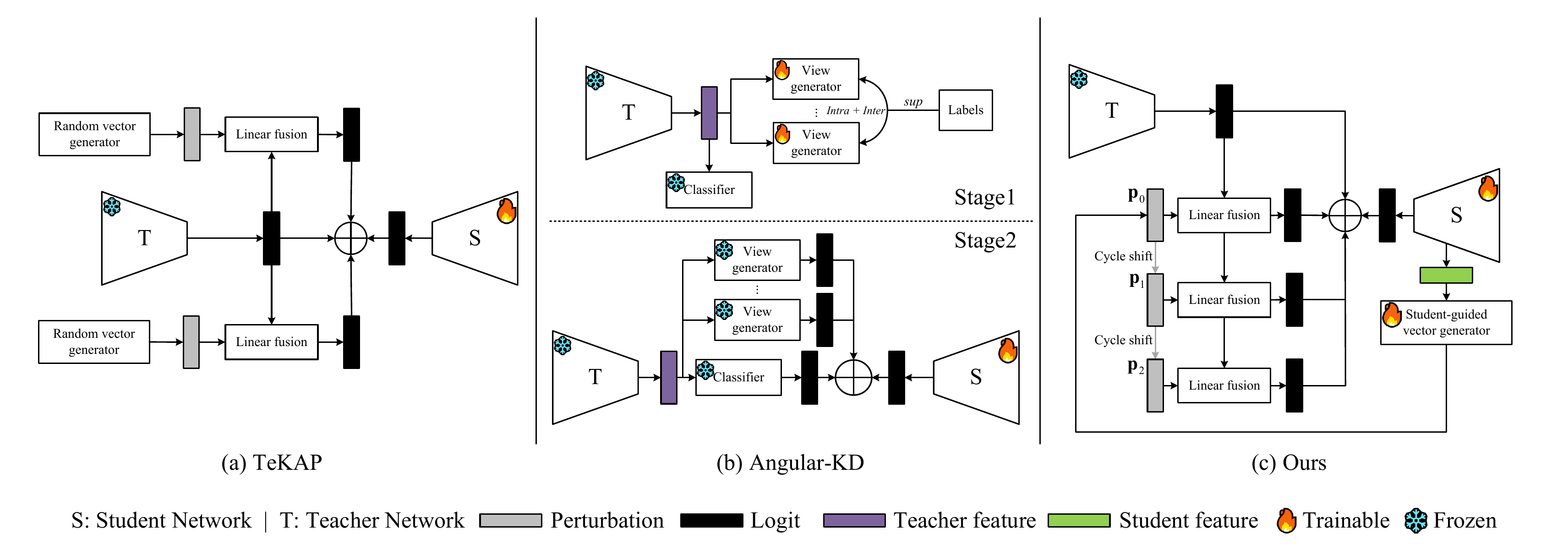}
    \caption{Training pipelines for distillation diversity augmentation. (a) TeKAP \cite{hossain2025single}: Single-stage with random noise addition. (b) Angular-KD \cite{yu2025single}: Two-stage training requiring pre-training of multiple generators using intra/inter-class losses and label supervision. (c) SAKD (Ours): Single-stage training that inherits TeKAP's augmentation philosophy but replaces random noise with learnable, student-guided perturbations.}
    \label{fig:pipeline}
\end{figure*}

Multi-teacher distillation aggregates knowledge from multiple independently trained models to mitigate this issue \cite{fukuda2017efficient,mirzadeh2020improved,son2021densely,multi-tea}. However, the cost of training and storing multiple large teachers is often prohibitive. A more practical direction generates multiple virtual teacher perspectives from a single pre-trained model. Existing approaches fall into two categories. Methods like TeKAP \cite{hossain2025single} (Fig.~\ref{fig:pipeline}(a)) inject random noise into teacher features or logits during training, creating diversity through controlled stochasticity in a single-stage, parameter-efficient manner.  TeKAP's key insight is to augment rather than replace the teacher's output, preserving the original knowledge as an anchor. While efficient, the noise in TeKAP is purely random and lacks semantic grounding. In contrast, structured methods like Angular-KD \cite{yu2025single} (Fig.~\ref{fig:pipeline}(b)) take a fundamentally different approach: they attach multiple trainable branches to the teacher, each tasked with reconstructing a complete, independent logit distribution from teacher features. This reconstruction task is inherently challenging and requires pre-training each branch on teacher features, necessitating a two-stage process where parameters scale linearly with the number of views.

We propose to combine the strengths of both approaches while addressing their limitations. We retain the elegant and efficient augmentation-based philosophy of TeKAP, where virtual teachers are formed by perturbing the original teacher output: $\mathbf{z}_i^T = \alpha \mathbf{z}^T + (1-\alpha) \mathbf{p}_i$. This formulation establishes a clear separation: the teacher's original logit $\mathbf{z}^T$ serves as the primary knowledge source, ensuring semantic correctness, while the perturbation $\mathbf{p}_i$ provides diversity enhancement. This design embodies two key advantages over reconstruction-based methods like Angular-KD. First, it follows a lightweight guidance principle: the generator only needs to learn structured perturbations, not precisely reconstruct the entire logit distribution, a significantly simpler task. Second, it ensures training stability: even if the perturbation $\mathbf{p}_i$ is imperfect, the teacher's original knowledge dominates (e.g., $\alpha = 0.9$), preventing the student from assimilating erroneous information and creating a stable learning anchor.

Building on this augmentation-based design, we are able to employ the weaker yet dynamic student features to generate $\mathbf{p}_i$, rather than relying on the stronger but static teacher features. This choice stems from the fact that the teacher's original output already guarantees semantic correctness; thus, perturbations only need to provide complementary diversity rather than reconstruct authoritative knowledge, reducing the reliance on the teacher. Utilizing student features shifts the student's role from a passive imitator to an active participant in its own training.
This student-guided mechanism creates a co-evolution loop where the student learns both from the teacher and how to enhance the teacher's supervision. More importantly, this design naturally enables single-stage end-to-end training, eliminating the requirement for separate pre-training of multiple generators on teacher features as in Angular-KD. Consequently, it resolves the fundamental tension between diversity and efficiency in teacher augmentation.

This insight leads to a surprisingly simple yet effective solution. Based on this insight, we introduce Shift-Augmented Knowledge  Distillation (SAKD). As shown in Fig.~\ref{fig:pipeline}(c), SAKD uses a single lightweight generator conditioned on the student's features to produce a base perturbation vector $\mathbf{p}_0$. SAKD retains TeKAP’s efficient augmentation framework but replaces random noise with adaptive, semantically guided perturbations. The perturbation $\mathbf{p}_0$ is then expanded into multiple distinct views through a parameter-free cyclic shift operation, maintaining constant parameter complexity regardless of the number of views generated. 


Our contributions are as follows:
\begin{itemize}
\item We introduce SAKD, a distillation framework that inherits the efficiency of augmentation-based methods but replaces random noise with learnable, student-guided perturbations. This design ensures training stability by treating teacher knowledge as the primary source and perturbations as complementary.
\item We propose a cyclic shift mechanism to efficiently create multiple complementary views from a single base perturbation, achieving constant parameter complexity and eliminating the need for multi-stage training.
\item Extensive experiments on CIFAR-100 and ImageNet demonstrate that SAKD consistently outperforms single-stage methods while achieving competitive accuracy with two-stage reconstruction-based methods.
\end{itemize}

\section{Method}
\label{sec:method}

\subsection{Preliminaries and Design Philosophy}
Given an input image $\mathbf{x}$, a pre-trained teacher network $\phi_T$ produces logit output $\mathbf{z}^T = \phi_T(\mathbf{x}) \in \mathbb{R}^C$ and intermediate features $\mathbf{f}^T \in \mathbb{R}^{d_T}$, where $C$ is the number of classes. The student network $\phi_S$ extracts intermediate feature representations $\mathbf{f}^S \in \mathbb{R}^{d_S}$ from a designated layer and produces prediction $\mathbf{z}^S = \phi_S(\mathbf{x}) \in \mathbb{R}^C$. Standard knowledge distillation minimizes:
\begin{equation}
\mathcal{L}_{\text{KD}} = \tau^2 \cdot \text{KL}\left(\sigma(\mathbf{z}^S/\tau) \,\|\, \sigma(\mathbf{z}^T/\tau)\right)
\end{equation}
where $\tau > 0$ is a temperature hyperparameter, and $\sigma(\cdot)$ denotes the softmax function.

Our design philosophy diverges from reconstruction-based approaches that require generators to produce complete logit distributions. Instead, we follow the augmentation paradigm, where diversity is introduced through perturbations to existing teacher knowledge. We formulate virtual teacher views as:
\begin{equation}
\mathbf{z}_i^T = \alpha \mathbf{z}^T + (1-\alpha) \mathbf{p}_i
\label{eq:augmentation}
\end{equation}
where $\alpha \in [0,1]$ controls the mixing ratio, $\mathbf{z}^T$ is the original teacher logit, and $\mathbf{p}_i \in \mathbb{R}^C$ is the generated perturbation. Our approach is inspired by TeKAP's augmentation paradigm, which creates diversity by adding random noise to the teacher's output. We retain this efficient framework but replace random noise with learnable structured perturbations. This formulation provides two main advantages. First, it follows a lightweight principle: the perturbation generator only needs to learn structured variations rather than precisely reconstructing the entire logit distribution, significantly reducing the learning complexity. Second, it ensures training stability: even when perturbations are suboptimal, the teacher's original knowledge remains dominant, preventing the student from learning incorrect information.

Leveraging these advantages, our method eliminates the need for teacher feature pre-training required by methods like Angular-KD, allowing less capable student features to contribute directly to teacher augmentation.

\subsection{Student-Guided Perturbation Generation}
Unlike view generators in reconstruction-based approaches that attempt to reconstruct complete teacher logits from features, our perturbation generator $\mathcal{G}$ serves a fundamentally different purpose: it learns only to produce complementary variations around established knowledge. This lightweight design conditions on student features $\mathbf{f}^S$, which provide a dynamic signal that adapts to the student's learning progress, while the teacher's original logit $\mathbf{z}^T$ remains the primary knowledge anchor. As a result, our generator only needs to learn perturbations rather than complete logits, making student features a suitable and efficient conditioning source that enables single-stage training without teacher feature pre-training.

Crucially, we guide the perturbation generation using student intermediate features, transforming the student's role from passive imitator to active participant. This approach naturally enables single-stage training, avoiding the two-stage pre-training requirement of methods like Angular-KD that rely on teacher features.

The generator is implemented as a lightweight MLP conditioned on student features, with dropout regularization:
\begin{equation}
\mathbf{p}_0 = \mathcal{G}(\mathbf{f}^S; \theta_G) = \mathbf{W}_2 \cdot \text{Dropout}(\mathbf{W}_1 \mathbf{f}^S + \mathbf{b}_1) + \mathbf{b}_2
\end{equation}
where $\text{Dropout}(\cdot)$ is applied element-wise with a fixed drop probability $p$ (set to $0.2$ in our experiments) and is active only during training. The parameters of the generator are $\theta_G = \{\mathbf{W}_1 \in \mathbb{R}^{d_h \times d_S}, \mathbf{b}_1 \in \mathbb{R}^{d_h}, \mathbf{W}_2 \in \mathbb{R}^{C \times d_h}, \mathbf{b}_2 \in \mathbb{R}^{C}\}$, with hidden dimension $d_h = 256$ by default. The conditioning on $\mathbf{f}^S$ creates an adaptive mechanism: as the student evolves throughout training, the generator's input becomes progressively more informative, enabling the perturbations to co-adapt with the student's learning state. This stands in contrast to methods that rely on static teacher features, where the conditioning signal remains fixed regardless of the student's progress.

\subsection{Cyclic Shift Expansion for Multi-View Generation}
To generate $N$ distinct virtual teacher views without the linear parameter overhead of maintaining multiple independent generators, we employ systematic cyclic shift operations. This design provides inherent initial diversity through structured transformations while maintaining constant parameter complexity, in contrast to multi-generator approaches where parameters scale linearly as $N \cdot |\theta_{\mathcal{G}}|$.

From the base perturbation $\mathbf{p}_0$, we generate $N$ distinct views through cyclic shifting:
\begin{equation}
\mathbf{p}_i = \text{CyclicShift}(\mathbf{p}_0, \delta_i), \quad \delta_i = i \cdot \Delta, \quad \Delta = \lfloor C/N \rfloor
\end{equation}
where for vector $\mathbf{v} = [v_1, v_2, \dots, v_C]^\top$, the cyclic shift operation rotates the elements by $\delta$ positions:
\begin{equation}
\text{CyclicShift}(\mathbf{v}, \delta) = [v_{k+1}, \dots, v_C, v_1, \dots, v_k]^\top, \quad k = C-\delta
\end{equation}
The cyclic shift provides a simple way to generate initial diversity by permuting dimensions of the base perturbation. The consistency loss and diversity loss introduced in the following section then jointly ensure that the shifted perturbations remain faithful to the teacher while being sufficiently distinct from each other. This design does not require the shifted views to be semantically precise individually, as the teacher's original logits are always preserved through the dominant mixing coefficient $\alpha$ in Eq.~\eqref{eq:augmentation}, providing a stable anchor throughout training. Importantly, the cyclic shift is a norm-preserving transformation: it keeps the $\ell_2$-norm of the perturbation unchanged ($\|\mathbf{p}_i\|_2 = \|\mathbf{p}_0\|_2$). This property ensures that the shift operation does not unintentionally amplify or suppress the magnitude of perturbations, allowing the consistency and diversity losses to control the perturbation characteristics in a stable manner. The shift also maintains constant parameter complexity regardless of the number of views.

\subsection{Optimization Framework}
We employ a dual-loss mechanism, conceptually similar to Angular-KD's intra-class and inter-class constraints but operating on lightweight perturbations, to ensure the generated perturbations are both semantically meaningful and complementary. This design maintains the integrity of teacher knowledge while encouraging diversity among perturbations for enriched supervision.

\noindent \textbf{Consistency Loss.} The consistency loss $\mathcal{L}_{\text{align}}$ grounds each perturbation in teacher knowledge, preventing arbitrary deviations. For each perturbation $\mathbf{p}_i \in \mathbb{R}^C$, we minimize its KL divergence from the original teacher logit $\mathbf{z}^T \in \mathbb{R}^C$:
\begin{equation}
\mathcal{L}_{\text{align}} = \sum_{i=1}^N \text{KL}\left(\sigma(\mathbf{p}_i/\tau) \,\|\, \sigma(\mathbf{z}^T/\tau)\right)
\end{equation}
A key distinction of this design is that we directly supervise the perturbations $\mathbf{p}_i$ rather than the final synthesized views. This makes the perturbation learning objective independent of the mixing coefficient $\alpha$, allowing $\alpha$ to serve as a tunable knob that controls augmentation strength without affecting how perturbations are learned.

\noindent \textbf{Diversity Loss.} The diversity loss $\mathcal{L}_{\text{div}}$ promotes complementarity among the $N$ views by minimizing pairwise cosine similarity. This encourages each view to provide unique information by penalizing redundant perturbations:
\begin{equation}
\mathcal{L}_{\text{div}} = \frac{1}{N(N-1)}\sum_{i=1}^N\sum_{j\neq i} \frac{\langle \mathbf{p}_i, \mathbf{p}_j \rangle}{\|\mathbf{p}_i\|\|\mathbf{p}_j\|}
\end{equation}
where $\langle \cdot, \cdot \rangle$ denotes the inner product. By pushing different perturbations away from each other, we ensure diverse supervisory signals for the student.

\noindent \textbf{Virtual Teacher Distillation Loss.} The virtual teacher distillation loss $\mathcal{L}_{\text{VD}}$ aligns student predictions $\mathbf{z}^S \in \mathbb{R}^C$ with the $N$ augmented teacher outputs $\mathbf{z}_i^T$:
\begin{equation}
\mathcal{L}_{\text{VD}} = \frac{1}{N}\sum_{i=1}^N \tau^2 \cdot \text{KL}\left(\sigma(\mathbf{z}^S/\tau) \,\|\, \sigma(\mathbf{z}_i^T/\tau)\right)
\end{equation}
This allows the student to learn from diverse yet consistent teacher variations, enhancing generalization.

\noindent \textbf{Complete Objective.} The complete objective integrates these components with the standard KD loss:
\begin{equation}
\mathcal{L}_{\text{total}} = \mathcal{L}_{\text{KD}} + \lambda_1\mathcal{L}_{\text{VD}} + \lambda_2\mathcal{L}_{\text{align}} + \lambda_3\mathcal{L}_{\text{div}}
\label{eq:total_loss}
\end{equation}
where $\lambda_1$, $\lambda_2$, and $\lambda_3$ are balancing weights. Following the practice of Angular-KD, we empirically set $\lambda_1=0.8$, $\lambda_2=1.0$, and $\lambda_3=1.0$ to balance the loss magnitudes. These values ensure that the virtual teacher distillation contributes significantly while maintaining the quality and diversity of generated perturbations.

\noindent \textbf{Adaptation to Feature Distillation.} For feature distillation scenarios, the framework naturally adapts by replacing logits $\mathbf{z}^T$ and $\mathbf{z}^S$ with corresponding teacher and student features $\mathbf{f}^T \in \mathbb{R}^{d_T}$ and $\mathbf{f}^S \in \mathbb{R}^{d_S}$, adjusting perturbation dimensions accordingly.

\subsection{Design Analysis and Justification}
\label{sec:design_analysis}

\noindent \textbf{Student Conditioning Unifies Training Stages.} 
Angular-KD \cite{yu2025single} relies on teacher features $\mathbf{f}^T$ as the conditioning signal for view generation. While these features naturally vary across input samples, the generator is trained during a separate warm-up stage. Once pre-training is complete, the generator's mapping from features to perturbations is fixed and does not adjust during the subsequent distillation stage. This separation is necessary because the teacher features provide no gradient pathway through which the student's learning state can influence the generator.

Our approach substitutes teacher features with student features $\mathbf{f}^S$ as the conditioning signal. This substitution fundamentally alters the optimization dynamics. Consider the gradients of the perturbation learning objectives $\mathcal{L}_{\text{align}}$ and $\mathcal{L}_{\text{div}}$ with respect to the student parameters $\theta_S$:
\begin{equation}
\frac{\partial \mathcal{L}_{\text{align}}}{\partial \theta_S} = 
\frac{\partial \mathcal{L}_{\text{align}}}{\partial \mathbf{p}} \frac{\partial \mathbf{p}}{\partial \mathbf{f}^S} \frac{\partial \mathbf{f}^S}{\partial \theta_S}, \quad
\frac{\partial \mathcal{L}_{\text{div}}}{\partial \theta_S} = 
\frac{\partial \mathcal{L}_{\text{div}}}{\partial \mathbf{p}} \frac{\partial \mathbf{p}}{\partial \mathbf{f}^S} \frac{\partial \mathbf{f}^S}{\partial \theta_S}
\end{equation}
These gradients enable the student backbone to receive supervisory signals derived from the perturbation generation process. In Angular-KD, the corresponding gradients vanish because $\partial \mathbf{f}^T / \partial \theta_S = 0$, making such feedback impossible.

This gradient connectivity has two implications. First, since the student features $\mathbf{f}^S$ evolve throughout training, the generator receives a progressively improving input without requiring separate pre-training. Second, the bidirectional interaction between the generator and the student allows both components to be optimized jointly from the start. Consequently, SAKD eliminates the warm-up stage entirely, unifying generator learning and student distillation into a single end-to-end training process.

\noindent \textbf{Discussion on Diversity Guarantee.}
The cyclic shift operation provides initial diversity through dimension permutation, while $\mathcal{L}_{\text{align}}$ and $\mathcal{L}_{\text{div}}$ refine the shifted perturbations to be semantically faithful and complementary. This two-stage refinement, consisting of structured initialization followed by loss-guided optimization, ensures that the final views provide diverse yet grounded supervision.  A formal analysis of how this combined mechanism reduces the upper bound of the ensemble expected loss follows the theoretical framework of Angular-KD \cite{yu2025single}.

\section{Experiments}
\label{sec:experiments}
\subsection{Experimental Settings}
We evaluate on CIFAR-100 \cite{krizhevsky2009learning} and ImageNet \cite{deng2009imagenet}. For CIFAR-100, we test the following teacher-student pairs:

\textbf{Similar architectures}: R32x4$\rightarrow$R8x4 (ResNet variants), W40-2$\rightarrow$W40-1 (Wide ResNets), VGG13$\rightarrow$VGG8 (VGG variants)

\textbf{Cross-architecture}: R32x4$\rightarrow$W40-2, R32x4$\rightarrow$W16-2 (Res-Net to Wide ResNet), W40-2$\rightarrow$R8x4 (Wide ResNet to ResNet)\\
where R$d$x$w$ denotes ResNet-$d$ with width multiplier $w$, W$d$-$w$ denotes Wide ResNet with depth $d$ and width factor $w$. For ImageNet, we use ResNet-34$\rightarrow$ResNet-18. 

Baselines include standard KD \cite{hinton2015distilling}, CRD \cite{tian2019contrastive}, DKD \cite{zhao2022decoupled}, MLKD \cite{jin2023multi}, the single-stage TeKAP (ICLR 2025) \cite{hossain2025single}, and the two-stage Angular-KD (NeurIPS 2025) \cite{yu2025single}. SAKD uses $N=3$ views, $\alpha=0.9$, $\tau=4.0$, $\lambda_1=0.8$, $\lambda_2=\lambda_3=1.0$. 

\noindent \textbf{Training details:} On CIFAR-100, models are trained for 240 epochs with SGD (momentum 0.9, weight decay $5\times10^{-4}$), batch size 64, initial learning rate 0.05, decayed by 0.1 at epochs 150, 180, and 210. On ImageNet, models are trained for 100 epochs with SGD (momentum 0.9, weight decay $1\times10^{-4}$), batch size 512, initial learning rate 0.2, decayed by 0.1 at epochs 30, 60, and 90. 

\subsection{Ablation Study}

We conduct ablation studies on CIFAR-100 (ResNet32×4 $\rightarrow$ ResNet8×4) to validate key design choices. Results in Table~\ref{tab:ablation} analyze three aspects: view generation strategy, mixing coefficient $\alpha$, and number of views $N$. Additionally, we evaluate the parameter and training efficiency of SAKD compared to Angular-KD in Table~\ref{tab:efficiency}.

\noindent \textbf{View Generation Strategy}: To evaluate the effectiveness of our cyclic shift strategy, we compare against three alternative view generation approaches. The first is a multi-generator baseline that trains separate MLPs for each view (analogous to Angular-KD's branch design but conditioned on student features for fair comparison), which achieves 75.42\% accuracy. The second is a no-augmentation baseline where all views are identical copies of the base perturbation $\mathbf{p}_0$, yielding only 74.85\% and confirming that diversity is essential for improvement. The third is a random permutation strategy where dimensions are independently shuffled for each view, attaining 75.16\%---outperforming no augmentation but still inferior to our cyclic shift. Our single-generator with cyclic shifts achieves the best accuracy of 75.54\%, demonstrating that the structured dimension permutation from cyclic shifts provides more effective and consistent diversity than random shuffling, while maintaining the parameter efficiency of a single generator.

\begin{table}[t]
\centering
\small
\caption{Ablation study of key design choices in SAKD on CIFAR-100 (ResNet32×4 $\rightarrow$ ResNet8×4).}
\setlength{\tabcolsep}{1pt}
\label{tab:ablation}
\begin{tabular}{lc}
\toprule
Component/Variant & Top-1 Accuracy (\%) \\
\midrule
\textit{View Generation Strategy} & \\
\hspace{1mm} Multiple independent generators (N=3) & 75.42 \\
\hspace{1mm} Single generator + no augmentation & 74.85 \\
\hspace{1mm} Single generator + random permutation (N=3) & 75.16 \\
\hspace{1mm} Single generator + cyclic shifts (Ours) & \textbf{75.54} \\
\midrule
\textit{Mixing Coefficient $\alpha$ in Eq.~\eqref{eq:augmentation}} & \\
\hspace{1mm} $\alpha = 0.0$ (fully generated) & 73.20 \\
\hspace{1mm} $\alpha = 0.3$ & 73.81 \\
\hspace{1mm} $\alpha = 0.5$ & 74.35 \\
\hspace{1mm} $\alpha = 0.7$ & 75.12 \\
\hspace{1mm} $\alpha = 0.9$ (default) & \textbf{75.54} \\
\hspace{1mm} $\alpha = 1.0$ (original teacher only) & 73.33 \\
\midrule
\textit{Number of Views $N$} & \\
\hspace{1mm} $N = 2$ & 75.31 \\
\hspace{1mm} $N = 3$ & \textbf{75.54} \\
\hspace{1mm} $N = 5$ & 75.38 \\
\hspace{1mm} $N = 7$ & 75.49 \\
\bottomrule
\end{tabular}
\end{table}

\begin{table}[t]
\centering
\small
\setlength{\tabcolsep}{1pt} 
\caption{Efficiency comparison on CIFAR-100 (R32x4$\rightarrow$R8x4).}
\label{tab:efficiency}
\begin{tabular}{lccc}
\toprule
Method & Extra Params ($N=3$) & Training Time & Pre-training \\
\midrule
Angular-KD \cite{yu2025single} & 276K & 3.1 hours & Required \\
SAKD (Ours) & \textbf{46K}& \textbf{2.4 hours} & \textbf{Not required} \\
\bottomrule
\end{tabular}
\end{table}

\begin{table*}[t]
\centering
\small
\setlength{\tabcolsep}{1.5pt} 
\caption{Comparison of different distillation methods on CIFAR-100. Best results in each column are bold.}
\label{tab:main_results}
\begin{tabular}{lccccccc}
\toprule
 & & \multicolumn{3}{c}{Same Architecture} & \multicolumn{3}{c}{Different Architecture} \\
\cmidrule(lr){3-5} \cmidrule(lr){6-8}
Base Method & Augmentation & R32x4$\rightarrow$R8x4 & W40-2$\rightarrow$W40-1 & VGG13$\rightarrow$VGG8 & R32×4$\rightarrow$W40-2 & R32×4$\rightarrow$W16-2 & W40-2$\rightarrow$R8×4 \\
\midrule
Teacher & / & 79.42 & 75.61 & 74.64 & 79.42 & 79.42 & 75.61 \\
Student & / & 72.50 & 71.98 & 70.36 & 75.61 & 73.26 & 72.50 \\
\midrule
\multirow{3}{*}{Logit Distil. \cite{hinton2015distilling}} 
 & w/o aug & 73.33 & 73.54 & 72.98 & 77.70 & 74.70 & 73.97 \\
 & + TeKAP \cite{hossain2025single}  & 74.79 & 73.80 & 74.00 & \textbf{77.97} & 75.08 & 75.09 \\
 & + SAKD & \textbf{75.54} & \textbf{74.02} & \textbf{74.34} & 77.89 & \textbf{75.33} & \textbf{76.25} \\
\midrule
\multirow{3}{*}{Feat. Distil. \cite{tian2019contrastive}} 
 & w/o aug & 75.51 & 74.14 & 73.94 & 78.15 & 75.65 & 75.24 \\
 & + TeKAP \cite{hossain2025single} & 75.65 & 74.21 & 74.10 & 78.05 & 75.42 & 75.65 \\
 & + SAKD & \textbf{75.79} & \textbf{74.52} & \textbf{74.36} & \textbf{78.31} & \textbf{75.88} & \textbf{76.22} \\
\midrule
\multirow{3}{*}{{Combined Distil. }} 
 & w/o aug & 75.46 & 74.38 & 74.29 & 78.01 & 75.86 & 75.57 \\
 & + TeKAP \cite{hossain2025single}& 75.98 & 74.38 & 74.42 & 78.68 & 75.62 & 75.90 \\
 & + SAKD & \textbf{76.48} & \textbf{74.94} & \textbf{75.01} & \textbf{78.79} & \textbf{76.05} & \textbf{76.55} \\
\bottomrule
\end{tabular}%
\end{table*}

\begin{table*}[t]
\centering
\small
\setlength{\tabcolsep}{1.5pt} 
\caption{Plug-and-Play Results on CIFAR-100. SAKD applied as an augmentation module to two state-of-the-art distillation methods: DKD and MLKD. Comparisons include the single-stage TeKAP and the two-stage Angular-KD. Best results are shown in \textbf{bold}, second-best are \underline{underlined}.}
\label{tab:plugplay}
\begin{tabular}{lccccccc}
\toprule
 & & \multicolumn{3}{c}{Same Architecture} & \multicolumn{3}{c}{Different Architecture} \\
\cmidrule(lr){3-5} \cmidrule(lr){6-8}
Base Method & Augmentation & R32x4$\rightarrow$R8x4 & VGG13$\rightarrow$VGG8 & W40-2$\rightarrow$W40-1 & R32x4$\rightarrow$W40-2 & R32x4$\rightarrow$W16-2 & W40-2$\rightarrow$R8x4 \\
\midrule
Teacher & / & 79.42 & 74.64 & 75.61 & 79.42 & 79.42 & 75.61 \\
Student & / & 72.50 & 70.36 & 71.98 & 75.61 & 73.26 & 72.50 \\
\midrule
\multirow{4}{*}{DKD~\cite{zhao2022decoupled}} & w/o aug & 76.32 & 74.68 & 74.81 & 78.46 & 75.70 & 75.56 \\
 & + TeKAP \cite{hossain2025single} & \textbf{76.65} & 74.55 & 73.83 & 78.64 & 75.28 & \uline{76.22} \\
   & + Angular-KD \cite{yu2025single} & 76.51 & \uline{74.76} & \textbf{74.89} & \textbf{78.99} & \uline{76.05} & 76.14 \\
 & + SAKD & \uline{76.61} & \textbf{74.94} & \uline{74.51} & \uline{78.86} & \textbf{76.10} & \textbf{76.26} \\
\midrule
\multirow{4}{*}{MLKD~\cite{jin2023multi}} & w/o aug & 77.08 & 75.18 & 75.35 & 79.26 & 76.52 & 77.33 \\
 & + TeKAP \cite{hossain2025single} & 77.04 & 75.37 & 75.31 & 78.72 & 76.46 & 77.28 \\
  & + Angular-KD \cite{yu2025single} & \uline{77.28} & \textbf{75.63} & \uline{75.37} & \uline{79.52} & \uline{76.60} & \uline{77.45} \\
 & + SAKD & \textbf{77.31} & \uline{75.45} & \textbf{75.40} & \textbf{79.71} & \textbf{76.68} & \textbf{77.61} \\
\bottomrule
\end{tabular}
\end{table*}

\noindent \textbf{Mixing Coefficient \(\alpha\):} The coefficient \(\alpha\) in Eq.~\eqref{eq:augmentation} governs the trade-off between the original teacher logits \(\mathbf{z}^T\) and the generated perturbations \(\mathbf{p}_i\). As shown in Table~\ref{tab:ablation}, performance is optimal at \(\alpha=0.9\) (75.54\%), indicating that teacher knowledge should remain dominant while perturbations supply beneficial diversity. At the extremes, performance degrades. Setting \(\alpha=0.0\) forces the virtual teachers to rely entirely on student-generated perturbations, which creates a degenerate learning loop: the generator learns to output perturbations that mimic the student's own logits, thereby trivially minimizing the distillation loss without transferring meaningful teacher knowledge. This leads to a collapse in generalization despite a low training loss. Conversely, \(\alpha=1.0\) (73.33\%) reduces to the original teacher-only supervision, offering no diversity and matching the performance of standard KD.

\noindent \textbf{Number of Views $N$}: The results show $N = 3$ achieves the highest accuracy (75.54\%), outperforming both $N=2$ (75.31\%) and $N=5$ (75.38\%). This may be because the cyclic shift operation with three views creates well-differentiated perturbations that offer optimal diversity. The minor performance drop at $N = 7$ (75.49\%) further indicates diminishing returns with excessive views.

\noindent\textbf{Efficiency Analysis}: As shown in Table~\ref{tab:efficiency}, SAKD demonstrates significant efficiency advantages over Angular-KD. With $N=3$ views, SAKD introduces only 46K extra parameters compared to Angular-KD's 276K, representing an 83\% reduction. This parameter efficiency remains constant regardless of $N$, whereas Angular-KD's parameters scale linearly with the number of views (460K for $N=5$). The efficiency translates directly to training time: SAKD completes in approximately 2.4 hours versus Angular-KD's 3.1 hours, while eliminating the pre-training stage entirely. This constant parameter complexity makes SAKD more scalable for resource-constrained applications.

\begin{figure*}[t]
    \centering
    \subfloat[KD Baseline]{\includegraphics[width=0.3\textwidth]{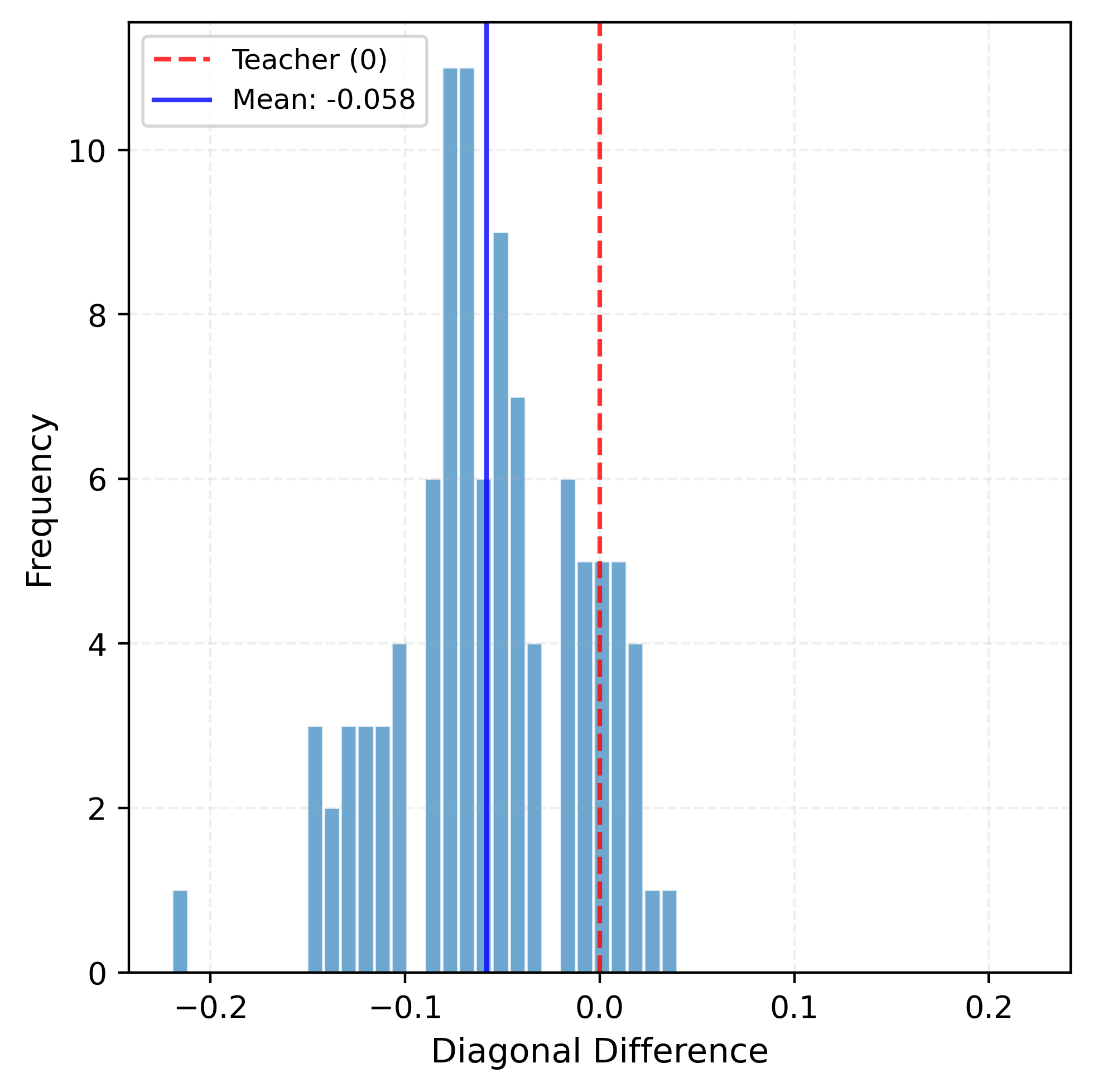}}
    \hfill
    \subfloat[TeKAP]{\includegraphics[width=0.3\textwidth]{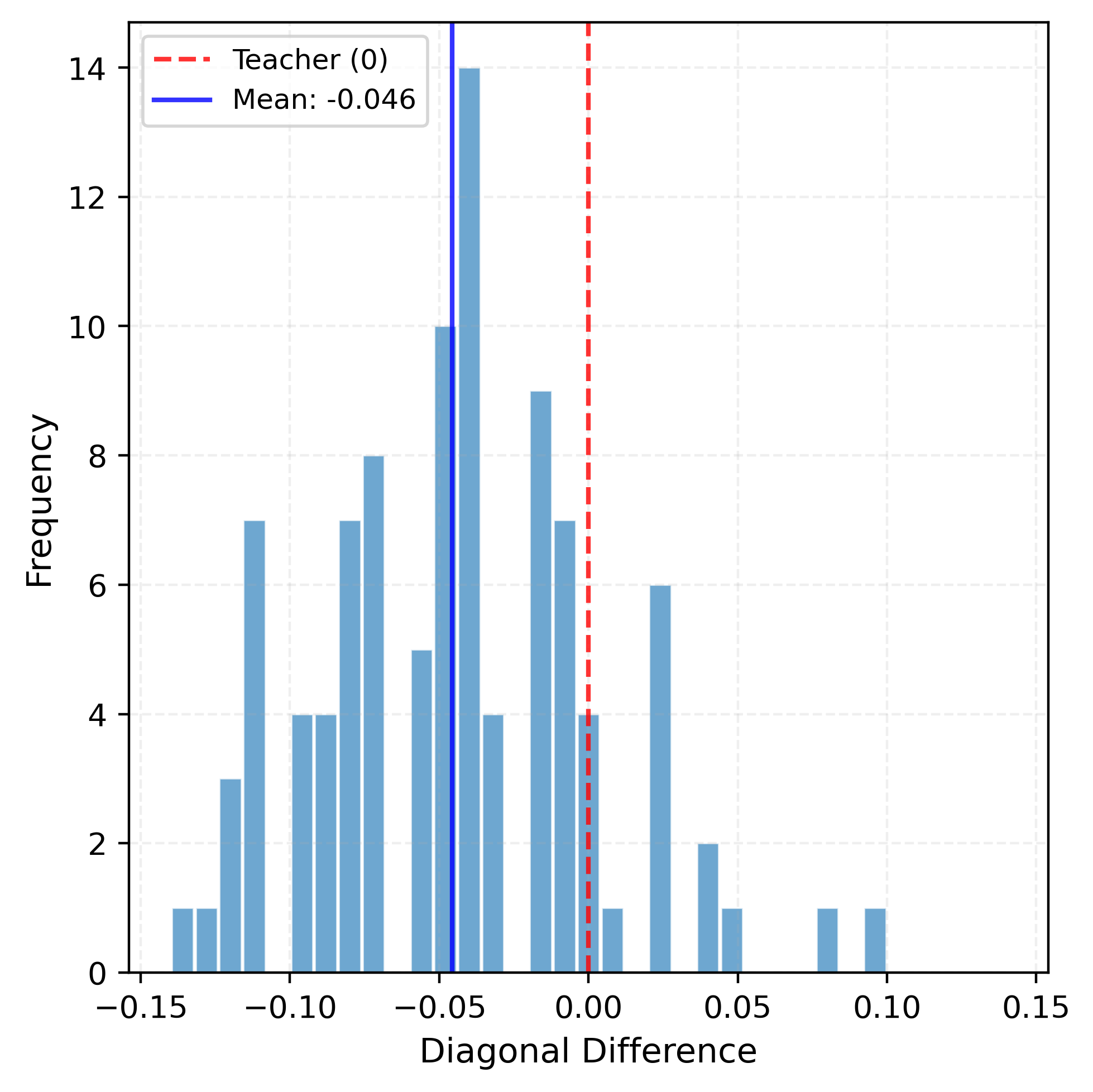}}
    \hfill
    \subfloat[SAKD (Ours)]{\includegraphics[width=0.3\textwidth]{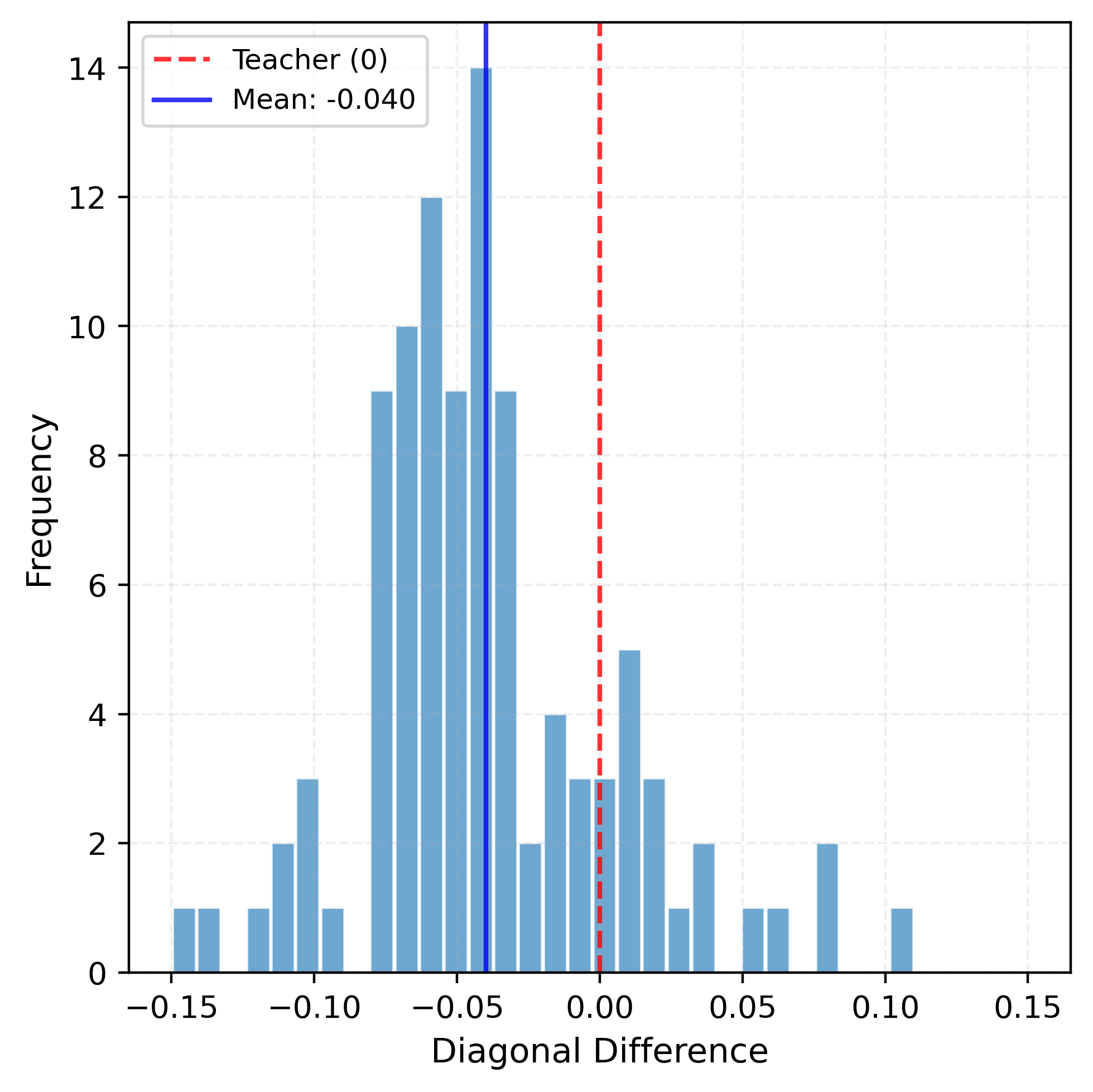}}
    \caption{Distribution of per-class diagonal differences between student and teacher normalized confusion matrices. SAKD shows the distribution most concentrated around zero (mean: -0.04), outperforming TeKAP (-0.046) and KD Baseline (-0.058).}
    \label{fig:diag_hist}
\end{figure*}
\subsection{Main Results on CIFAR-100}
\label{subsec:cifar100_results}
Table~\ref{tab:main_results} presents a comprehensive evaluation of SAKD on CIFAR-100 across six different teacher-student architecture pairs, first comparing it with TeKAP as both are single-stage augmentation approaches. SAKD consistently outperforms the random perturbation method TeKAP in both logit and feature distillation paradigms.
In logit distillation, SAKD achieves superior performance in 5 out of 6 architecture pairs. The improvement is particularly significant on the cross-architecture pair W40-2$\rightarrow$R8×4, where SAKD attains 76.25\% accuracy compared to TeKAP's 75.09\%, representing a 1.16\% absolute improvement. On the same architecture pair R32x4$\rightarrow$R8x4, SAKD reaches 75.54\% versus TeKAP's 74.79\%.
For feature distillation, SAKD maintains consistent advantages, achieving the highest accuracy in 4 out of 6 experimental settings. On W40-2$\rightarrow$W40-1, SAKD obtains 74.52\% compared to TeKAP's 74.21\%, and on R32×4$\rightarrow$W16-2, SAKD reaches 75.88\% versus TeKAP's 75.42\%.

When combining both logit and feature distillation strategies, SAKD's performance advantage becomes most pronounced. On R32x4$\rightarrow$R8x4, SAKD achieves 76.48\% accuracy, significantly surpassing TeKAP's 75.98\% by 0.50\%. More remarkably, on VGG13$\rightarrow$VGG8, SAKD attains 75.01\%, the highest among all compared methods and 0.59\% higher than TeKAP's 74.42\%. These results demonstrate that our structured perturbations generated through student guidance and cyclic shifts provide more effective diversity than TeKAP's random Gaussian noise, leading to superior knowledge transfer across various architecture configurations.

\subsection{Plug-and-Play Results on CIFAR-100}
\label{subsec:plugplay_results}
To further evaluate the compatibility and effectiveness of SAKD, we apply it as a plug-and-play augmentation module to two state-of-the-art distillation methods: DKD and MLKD. Results are presented in Table~\ref{tab:plugplay} (best in bold, second-best underlined). Compared with the single-stage TeKAP and the two-stage Angular-KD, SAKD demonstrates exceptional training stability and performance robustness.

The most salient advantage of SAKD is its consistent performance. As shown in the table, SAKD ranks either first or second in all 12 experimental settings (achieving the best accuracy in 8 cases and the second-best in the remaining 4). 
SAKD ranks among the top two in all 12 experimental settings, demonstrating the reliability of our student-guided perturbation design.
In contrast, while the two-stage Angular-KD achieves competitive results (ranking first in 3 cases), it shows noticeable instability: it falls to third place in two challenging pairs under the DKD baseline (R32x4$\rightarrow$R8x4 and W40-2$\rightarrow$R8x4).
When integrated with DKD, SAKD delivers strong and stable results. It attains the highest accuracy on VGG13$\rightarrow$VGG8 (74.94\%) and the second highest on R32x4$\rightarrow$R8x4 (76.61\%), in both cases surpassing Angular-KD. The advantage becomes even clearer under the more advanced MLKD baseline. Here, SAKD achieves the best accuracy in 5 out of 6 pairs, including a notable 0.19\% lead over Angular-KD on the challenging cross-architecture pair R32x4$\rightarrow$W40-2.

These results validate that our single-stage, student-guided framework can not only match but often exceed the performance of the more complex two-stage Angular-KD, while completely avoiding its costly pre-training phase. Moreover, SAKD consistently outperforms the other single-stage baseline, TeKAP, in 11 out of 12 settings, confirming that our structured, learnable perturbations provide more beneficial supervisory diversity than simple random noise.

In summary, SAKD successfully bridges the gap between efficiency and performance: it retains the training simplicity and parameter efficiency of single-stage methods while delivering accuracy and stability that rival or surpass those of two-stage methods. This makes SAKD a highly practical and effective choice for knowledge distillation in resource-constrained scenarios.

\begin{table}[htbp]
\centering
\small
\caption{Results on ImageNet. Logit-level augmentations are applied for both TeKAP and SAKD under a ResNet34 teacher and ResNet18 student.}
\label{tab:imagenet}
\begin{tabular}{lcc}
\toprule
Method & Top-1 Acc (\%) & Top-5 Acc (\%) \\
\midrule
Teacher (ResNet-34) & 73.31 & 91.42 \\
Student (ResNet-18) & 69.75 & 89.07 \\
\midrule
w/o aug & 70.41 & 89.70 \\
TeKAP \cite{hossain2025single} & 70.63 & 89.92 \\
Angular-KD \cite{yu2025single} & 71.07 & 90.39 \\
\textbf{SAKD (Ours)} & \textbf{71.12} & \textbf{90.41} \\
\bottomrule
\end{tabular}
\end{table}
\subsection{Scalability on ImageNet}
\label{subsec:imagenet_results}
We evaluate SAKD on ImageNet with ResNet-34$\rightarrow$ResNet-18. As shown in Table~\ref{tab:imagenet}, SAKD achieves 71.12\% top-1 and 90.41\% top-5 accuracy, outperforming TeKAP (70.63\% / 89.92\%) and Angular-KD (71.07\% / 90.39\%). The 0.49\% top-1 improvement over TeKAP demonstrates that structured perturbations are particularly beneficial for complex datasets. More importantly, SAKD surpasses Angular-KD in both top-1 and top-5 accuracy while requiring only a single training stage and eliminating costly pre-training. This validates that student-guided perturbations can effectively replace teacher-feature pre-training, offering a more efficient and competitive alternative for large-scale applications.

\begin{figure}[t]
    \centering
    \includegraphics[width=0.95\linewidth]{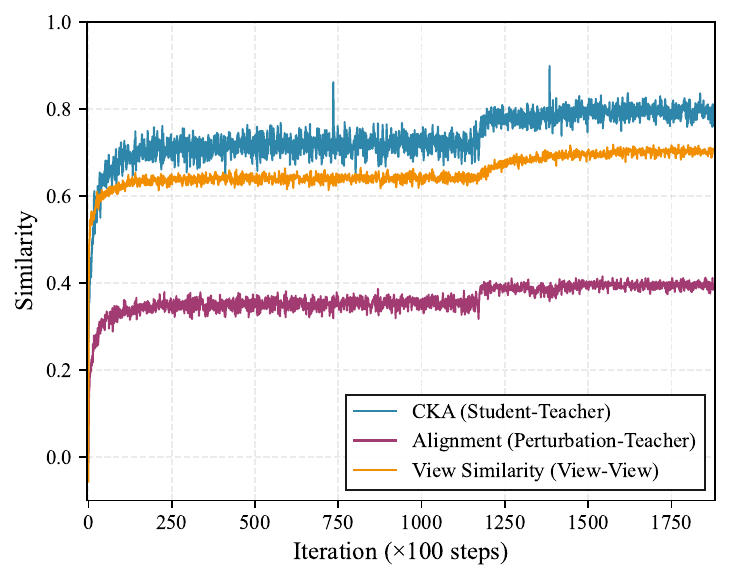}
    \caption{Training dynamics of CKA (student-teacher similarity), 
    perturbation alignment, and intra-view similarity on CIFAR-100 
    (ResNet32$\times$4 $\rightarrow$ ResNet8$\times$4).}
    \label{fig:perturbation_dynamics}
\end{figure}

\subsection{Analysis and Visualization}
\subsubsection{Dynamics of Student-Guided Perturbation Generation}
\label{subsec:perturbation_dynamics}

To validate the co-adaptation mechanism between the student and 
perturbation generator, we track three metrics throughout training 
on CIFAR-100 (ResNet32$\times$4 $\rightarrow$ ResNet8$\times$4): 
CKA similarity between student and teacher features, perturbation 
alignment (cosine similarity between perturbations and teacher logits), 
and intra-view similarity among generated views.

As shown in Fig.~\ref{fig:perturbation_dynamics}, CKA rises steadily 
throughout training, confirming that the student progressively aligns 
with the teacher's representational space. Perturbation alignment 
follows a similar upward trajectory, demonstrating that perturbation 
quality improves as the student features become more informative. 
The alignment remains moderate by design: the teacher's original logit, 
weighted by $\alpha=0.9$, provides the primary semantic supervision, 
while perturbations only supply complementary diversity. Meanwhile, 
intra-view similarity stabilizes at a healthy level, confirming that 
the generated views maintain diversity without collapsing to the 
teacher. These results provide direct evidence for the co-adaptation 
between the student and perturbation generator.

\subsubsection{Distribution of Per-Class Discrepancy}
\label{subsec:per_class_discrepancy}
To evaluate how consistently the student preserves the teacher's 
class-wise decision behavior, we compute the diagonal difference 
between their normalized confusion matrices:
\begin{equation}
\Delta_c = \text{diag}(\mathbf{C}_S - \mathbf{C}_T)_c
\end{equation}
where values close to zero indicate per-class consistency with 
the teacher. Fig.~\ref{fig:diag_hist} presents the frequency 
distribution of these differences across three methods. KD 
Baseline shows a broad distribution centered at -0.058, with 
substantial deviations in many classes. TeKAP improves this to 
-0.046 with a slightly narrower distribution. SAKD achieves 
the best result with mean -0.040 and the most concentrated 
distribution around zero, while exhibiting significantly fewer 
classes with large deviations beyond $\pm 0.1$. This demonstrates 
that student-guided perturbations enable the student to more 
uniformly preserve the teacher's decision logic across all classes.
\begin{figure}[t]
\centering
\includegraphics[width=\linewidth]{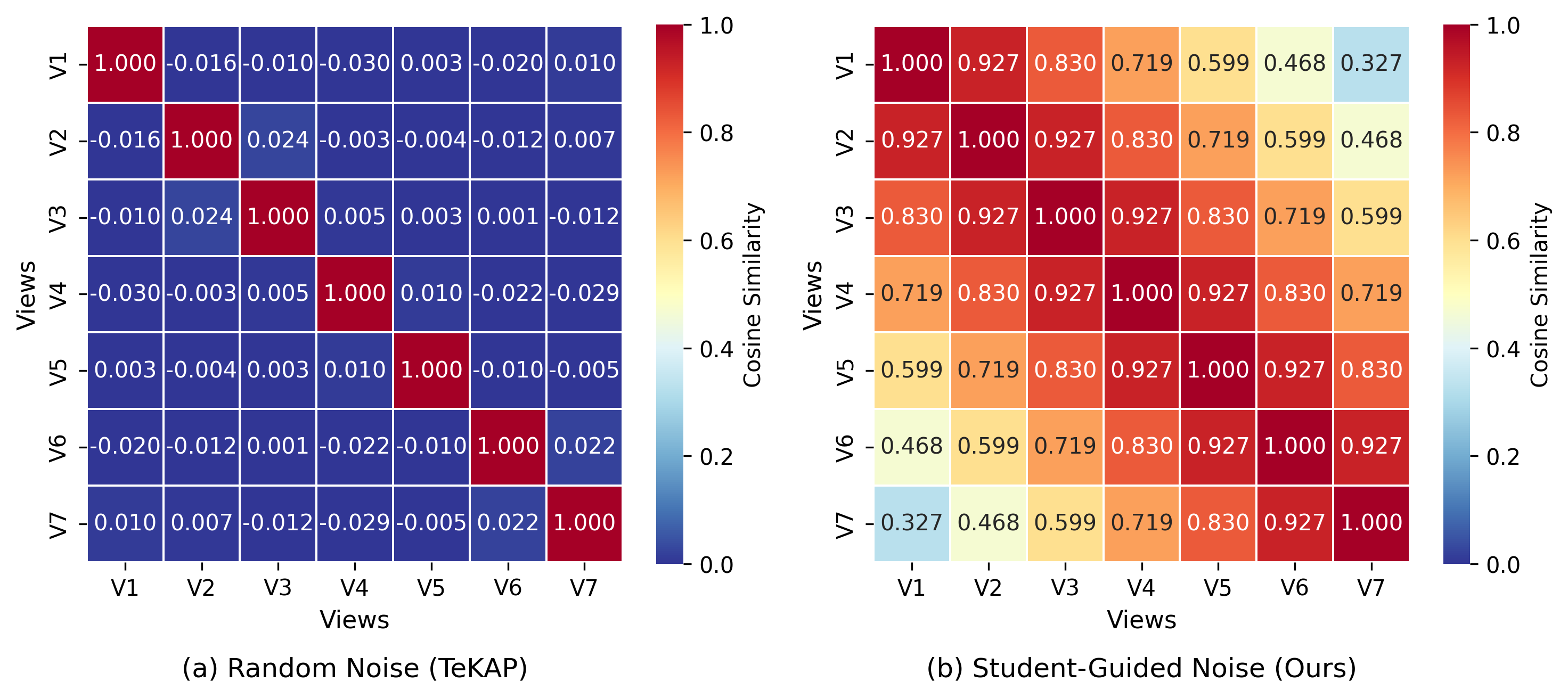}
\caption{Correlation matrices of perturbation vectors ($N=7$ as an illustrative example): (a) TeKAP: random noise with near-zero correlations; (b) SAKD: structured perturbation with distance-decay pattern.}
\label{fig:noise_corr}
\end{figure}
\subsubsection{Structured Perturbation Analysis}
Fig.~\ref{fig:noise_corr} compares correlation matrices between perturbed vectors $\mathbf{p}_i$ of TeKAP and SAKD. TeKAP's random noise shows near-zero correlations, indicating unstructured randomness. In contrast, SAKD's perturbations exhibit structured correlations with a clear distance-decay pattern: adjacent views are highly similar (correlation 0.927), while distant views show lower similarity (0.327). This structured diversity arises from our cyclic shift mechanism combined with semantic alignment constraints. The gradual variation in similarity across views provides a more natural and informative supervisory signal than random noise. Neighboring views reinforce core concepts, and more distant views introduce controlled diversity, offering multiple complementary perspectives on the teacher's knowledge that enhance the distillation process.
\section{Conclusion}
We present SAKD, a novel knowledge distillation framework that leverages student features to guide teacher knowledge augmentation. Unlike prior teacher-conditioned approaches that require separate pre-training stages, our method uses the student's evolving features as a dynamic conditioning signal for perturbation generation, unifying generator learning and student distillation into a single end-to-end training process. By generating adaptive perturbations through a student-guided generator and expanding them via parameter-free cyclic shifts, SAKD creates effective diversity through structured perturbations while maintaining single-stage trainability and constant parameter complexity. Extensive experiments on CIFAR-100 and ImageNet demonstrate that SAKD consistently outperforms existing augmentation methods in both accuracy and efficiency, achieving competitive performance with two-stage approaches while eliminating their costly pre-training requirements. This work presents an initial exploration in which cyclic shifts represent one viable transformation among many alternatives. Future work will include exploring alternative transformations, optimizing perturbation generation, and extending SAKD to vision tasks like detection and segmentation.

\bibliographystyle{ieeetr}
\bibliography{ref} 

\end{document}